\documentclass[twoside]{article}

\usepackage[accepted]{aistats2022}
\usepackage[round]{natbib}

\PassOptionsToPackage{round}{natbib}
\renewcommand{\cite}{\citep}

\usepackage[toc,page]{appendix}
\usepackage[utf8]{inputenc} 
\usepackage[T1]{fontenc}    
\usepackage{hyperref}       
\usepackage{url}            
\usepackage{booktabs}       
\usepackage{amsfonts}       
\usepackage{nicefrac}       
\usepackage{microtype}      
\usepackage{bm}
\usepackage{paralist}
\usepackage{mathtools}      
\usepackage{empheq}         
\usepackage{amsmath}        
\usepackage{amsthm}         
\newtheorem{Theorem.}{Theorem}
\newtheorem{Definition.}{Definition}
\newtheorem{Proposition.}{Proposition}
\newtheorem{Corollary.}{Corollary}
\usepackage{amssymb}        
\usepackage[ruled,vlined]{algorithm2e}
\usepackage{braket}                               
\usepackage{graphicx}                   
\newcommand{\argmin}{\mathop\textrm{argmin}\limits} 

\usepackage{thmtools}
\declaretheoremstyle[
  spaceabove=0pt, spacebelow=0pt,
  headfont=\itshape,
  notefont=,
  notebraces={(}{)},
  postheadspace=1em,
  numbered=no,
  qed=$\square$
]{myproof}
\declaretheorem[title=Proof, style=myproof]{myproof}
\renewenvironment{proof}{\begin{myproof}}{\end{myproof}}

\renewenvironment{itemize}{%
\begin{list}{\textbullet\ \ }{%
 \setlength{\itemindent}{0pt}
 \setlength{\leftmargin}{2em}%
 \setlength{\rightmargin}{0pt}%
 \setlength{\labelsep}{0pt}%
 \setlength{\labelwidth}{3em}%
 \setlength{\itemsep}{0em}%
 \setlength{\parsep}{0em}%
 \setlength{\listparindent}{0pt}%
 \setlength{\topsep}{0em}
}}{\end{list}}
\newcounter{enum2}

\begin{document}

%

%

\twocolumn[
\aistatstitle{Fast Rank-1 NMF for Missing Data with KL Divergence}

\aistatsauthor{ Kazu Ghalamkari \And Mahito Sugiyama}
\aistatsaddress{ 
National Institute of Informatics \\ 
The Graduate University for Advanced Studies, SOKENDAI
}
]

\begin{abstract}
We propose a fast non-gradient-based method of rank-1 non-negative matrix factorization (NMF) for missing data, called A1GM, that minimizes the KL divergence from an input matrix to the reconstructed rank-1 matrix. Our method is based on our new finding of an analytical closed-formula of the best rank-1 non-negative multiple matrix factorization (NMMF), a variety of NMF. NMMF is known to exactly solve NMF for missing data if positions of missing values satisfy a certain condition, and A1GM transforms a given matrix so that the analytical solution to NMMF can be applied. We empirically show that A1GM is more efficient than a gradient method with competitive reconstruction errors.
\end{abstract}

\section{INTRODUCTION}
The tabular form is one of the most common data types across different fields, and includes purchasing data, processed sensor data, and images. Tabular datasets are often treated as matrices for analysis.
To date, many methods have been developed to extract essential information from matrices~\cite{thompson1984canonical, wold1987principal, comon1994independent,blei2003latent}. Non-negative matrix factorization (NMF) is one of the most popular techniques~\cite{lee1999learning}. NMF extracts factors in a dataset by decomposing a given non-negative matrix $\mathbf{X}\in\mathbb{R}_{\geq 0}^{I \times J}$ into a product $\mathbf{AB}$ of two matrices $\mathbf{A}\in\mathbb{R}_{\geq 0}^{I \times r}$ and $\mathbf{B}\in\mathbb{R}_{\geq 0}^{r \times J}$ so that the predetermined cost becomes smaller. The matrix rank of the product $\mathbf{AB}$ is less than or equal to the hyperparameter $r \in \mathbb{N}$.

Standard NMF, which uses the least squares error $\|\mathbf{X-AB}\|_\mathrm{F}$ as the cost function, is widely used in various applications, including face recognition~\cite{rajapakse2004color}, recommender systems~\cite{takacs2008investigation}, and text analysis~\cite{xu2003document}. Although it is an NP-hard problem to find the best decomposition that minimizes the cost function exactly for any $r>1$~\cite{vavasis2010complexity}, in an exceptional case of $r = 1$, the best decomposition is obtained in polynomial time~\cite{edward2007}.
Therefore it is possible to obtain the best decomposition in a reasonable time if $r = 1$ and 
a number of NMF algorithms are developed based on rank-$1$ NMF~\cite{biggs2008nonnegative,liao2013efficient}. 
Rank-$1$ NMF approximates an input matrix $\mathbf{X}$ by the Kronecker product $\bm{a}\otimes\bm{b}$ of two non-negative vectors $\bm{a}$ and $\bm{b}$, called dominant factors. Since the vectors $\bm{a}$ and $\bm{b}$ correspond to the largest principal components restricted to the first quadrant in eigenvalue decomposition, they are representative features that roughly describe the input matrix.

However, it is problematic if a matrix includes missing values, which often occurs in real-world datasets in practice. When the cost function is the least squares error, NMF for missing data --- that is, a matrix with missing values, which we call \emph{missing NMF}~\cite{kim2009weighted} --- is known to become an NP-hard problem, even for $r = 1$~\cite{gillis2011low}. Although this is a crucial drawback of missing NMF, missing NMF with other cost functions has not been well studied to date. In this paper, we show that there are certain cases in which rank-$1$ missing NMF can be exactly solved in polynomial time when the cost function is defined as the KL divergence. 

Our key idea is that, instead of directly solving missing NMF, we focus on non-negative multiple matrix factorization (NMMF)~\cite{takeuchi2013non}, a variant of NMF.
We derive a closed formula that globally minimizes the cost function of NMMF when the target rank $r = 1$, which is our main theoretical contribution.
NMMF conducts simultaneous factor-sharing decomposition of multiple matrices, which has been used in purchase forecast systems~\cite{kohjima2016non} and recommender systems~\cite{zhang2016non}. Interestingly, if the cost function is given as the KL divergence, NMMF is equivalent to missing NMF when missing values are clustered in the lower right corner of the input matrix~\cite{takeuchi2013non}. 
Using this relationship between missing NMF and NMMF, 
we can efficiently compute the exact solution of rank-$1$ missing NMF with the KL divergence by our solution to NMMF when we can locate missing values in a rectangular region by permuting rows and columns. 

Moreover, to treat matrices in which we cannot locate missing values in a rectangular region by permutations of rows and columns, we provide a method of finding an approximate solution of rank-$1$ missing NMF by adding more missing values so that we can use the closed formula of the best rank-$1$ NMMF. We call our novel method \emph{A1GM} (Analytical solution for rank-1 NMF with Grid-based Missing values). We empirically show that A1GM is more efficient than an existing gradient-based method for rank-$1$ missing NMF with the competitive reconstruction error.

We summarize our contribution as follows:
\begin{itemize}
    \item We derive a closed formula of the best rank-$1$ NMMF, which extracts the most dominant factors faster than the existing gradient method.
    \item We develop a novel efficient method to solve rank-$1$ missing NMF, called A1GM, and prove that A1GM globally minimizes a cost function under an assumption about the position of missing values.
    \item We empirically show that A1GM is more efficient than an existing method for missing NMF with competitive reconstruction error.
\end{itemize}

\subsection{Notations and problem setup}
Matrices are denoted by bold capital letters like $\mathbf{X}$ and $\mathbf{Y}$, and vectors are denoted by lower-case bold alphabets like $\bm{a}$ and $\bm{b}$. The total sum of a matrix or a vector is represented as $S(\cdot)$. The $i$th component of a vector $\bm{a}$ is written in its non-bold letter as $a_i$. The Kronecker product of two vectors $\bm{a}$ and $\bm{b}$ is denoted by $(\bm{a} \otimes \bm{b})$, which is a rank-$1$ matrix, and each element is defined as $(\bm{a} \otimes \bm{b})_{ij}=a_ib_j$. The $I \times J$ all-one and all-zero matrices are denoted by $\mathbf{1}_{I \times J}$ and $\mathbf{0}_{I \times J}$, respectively. The identity matrix is denoted by $\mathbf{I}$. The transpose of a matrix $\mathbf{X}$ is denoted by $\mathbf{X}^\mathrm{T}$. The element-wise product of two matrices $\mathbf{A}$ and $\mathbf{B}$ is denoted by $\mathbf{A} \circ \mathbf{B}$. For a pair of natural numbers $n$ and $m$ $(\geq n)$, we denote by $[n,m]=\{n,n+1,\dots, m-1,m\}$. We abbreviate $[1,m]$ as $[m]$. The set difference of $B$ and $A$ is denoted by $B \setminus A$.
When we use the Kullback--Leibler (KL) divergence $D(\mathbf{X},\mathbf{Y})$ for matrices $\mathbf{X}$ and $\mathbf{Y}$, it is defined as follows:~\cite{NIPS2000_f9d11525}
    \begin{align*}
            {D(\mathbf{X}, \mathbf{Y})}
            = \sum_{i,j}
            \left\{ \mathbf{X}_{ij} \log{\frac{\mathbf{X}_{ij}}{\mathbf{Y}_{ij}}}
            -  \mathbf{X}_{ij}
            +  \mathbf{Y}_{ij}
            \right\}.
    \end{align*}
In this paper, we treat two tasks, rank-$1$ NMMF and rank-$1$ missing NMF. We consistently assume that these cost functions are defined by the above KL divergence throughout the paper. 
\paragraph{Rank-1 NMMF} 
Rank-1 NMMF simultaneously decomposes three matrices $\mathbf{X}\in\mathbb{R}_{\geq 0}^{I \times J}$, $\mathbf{Y}\in\mathbb{R}_{\geq 0}^{N \times J}$ and $\mathbf{Z}\in\mathbb{R}_{\geq 0}^{I \times M}$ into three rank-$1$ matrices $\bm{w}\otimes\bm{h}$, $\bm{a}\otimes\bm{h}$, and $\bm{w}\otimes\bm{b}$, respectively, using non-negative vectors $\bm{w}\in\mathbb{R}_{\geq 0}^{I}, \bm{h}\in\mathbb{R}_{\geq 0}^{ J}, \bm{a}\in\mathbb{R}_{\geq 0}^{N}$, and $\bm{b}\in\mathbb{R}_{\geq 0}^{M}$. The cost function of NMMF is defined as
    \begin{align}\label{eq:cost_NMMF}
        D(\mathbf{X},\bm{w}\otimes\bm{h})+\alpha D(\mathbf{Y},\bm{a}\otimes\bm{h})+\beta D(\mathbf{Z},\bm{w}\otimes\bm{b}),
    \end{align}
and the task of rank-1 NMMF is to find vectors $\bm{w}$, $\bm{h}$, $\bm{a}$, and $\bm{b}$ that minimize the above cost.
We assume that the scaling parameters $\alpha$ and $\beta$ are non-negative

\paragraph{Rank-1 missing NMF} 

Rank-1 NMF for a matrix with missing values (rank-1 missing NMF) is the task of finding two non-negative vectors $\bm{w}\in\mathbb{R}^{I}_{\geq 0}$ and $\bm{h}\in\mathbb{R}^{J}_{\geq 0}$ that minimize a weighted cost function $D_{\Phi}(\mathbf{X},\bm{w}\otimes\bm{h})$ defined as
\begin{align}\label{eq:cost_WNMF}
    D_{\mathbf{\Phi}}(\mathbf{X},\bm{w}\otimes\bm{h})=D(\mathbf{\Phi}\circ \mathbf{X}, \mathbf{\Phi}\circ \left(\bm{w}\otimes\bm{h}\right))
\end{align}
for a non-negative input matrix $\mathbf{X}\in\mathbb{R}_{\geq 0}^{I \times J}$ and a binary weight matrix $\mathbf{\Phi}\in\{0,1\}^{I \times J}$. The weight matrix indicates the position of missing values; that is, $\mathbf{\Phi}_{ij}=0$ if the entry $\mathbf{X}_{ij}$ is missing, $\mathbf{\Phi}_{ij}=1$ otherwise. Note that the above cost function \eqref{eq:cost_WNMF} is always convex in $\bm{w}$ and $\bm{h}$.
\section{THE A1GM METHOD}
In this section, we present our proposed method for rank-$1$ missing NMF, called A1GM. We derive a closed formula of the best rank-$1$ NMMF in Section~\ref{subsec:NMMF}, followed by introducing A1GM using the closed formula in Section~\ref{subsec:A1GM}.

\subsection{A closed formula of the best rank-1 NMMF}\label{subsec:NMMF}
We give the following closed formula of the best rank-$1$ NMMF that exactly minimizes the cost function in Equation~\eqref{eq:cost_NMMF}, which is our main theoretical contribution. This formula efficiently extracts only the most dominant factors from three input matrices.
\begin{Theorem.}[the best rank-$1$ NMMF]\label{th:best_rank1_NMMF}
        For any given three positive matrices $\mathbf{X} \in \mathbb{R}^{I\times J}_{> 0}$, $\mathbf{Y}\in\mathbb{R}_{> 0}^{N \times J}$, and $\mathbf{Z}\in\mathbb{R}_{> 0}^{I \times M}$ and two parameters $\alpha, \beta \ge 0$, four non-negative vectors $\bm{w}\in\mathbb{R}_{\geq 0}^{I}, \bm{h}\in\mathbb{R}_{\geq 0}^{ J}, \bm{a}\in\mathbb{R}_{\geq 0}^{N}$, and $\bm{b}\in\mathbb{R}_{\geq 0}^{M}$ that minimize the cost function in Equation~\eqref{eq:cost_NMMF} is given as 
         \begin{align*}
            w_i&=\frac{\sqrt{S(\mathbf{X})}}{S(\mathbf{X})+\beta S(\mathbf{Z})}
            \left\{\sum_{j=1}^J\mathbf{X}_{ij}+\sum_{m=1}^M\beta \mathbf{Z}_{im}\right\}, \\
            h_j&=\frac{\sqrt{S(\mathbf{X})}}{S(\mathbf{X})+\alpha S(\mathbf{Y})}
            \left\{\sum_{i=1}^I\mathbf{X}_{ij}+\sum_{n=1}^N\alpha \mathbf{Y}_{nj}\right\}, \\
            a_{n}&=\frac{\sum_{j=1}^J \mathbf{Y}_{nj}}{\sqrt{S(\mathbf{X})}}, \quad 
            b_{m}=\frac{\sum_{i=1}^I \mathbf{Z}_{im}}{\sqrt{S(\mathbf{X})}}.
         \end{align*}
    \end{Theorem.}
The time complexity to obtain the best rank-1 NMMF is $O(IJ+NJ+IM)$.
Note that, for $N=M=0$, our result in Theorem~\ref{th:best_rank1_NMMF} coincides with the best rank-$1$ NMF minimizing the KL divergence from an input matrix $\mathbf{X}$ to the reconstructed rank-$1$ matrix shown in~\citet{ho2008non}.

In subsections \ref{subsec:proj} -- \ref{subsec:the_best_NMMF}, we introduce information geometric formulation of NMMF using the log-linear model and derive the closed-form solution given in Theorem~\ref{th:best_rank1_NMMF}.
Although they can be viewed as a proof of Theorem~\ref{th:best_rank1_NMMF}, we provide it in the main text as it includes various interesting properties between NMMF and information geometry, particularly characterization of rank-$1$ decomposition via parameters of the exponential family, which we believe is valuable for further development of this line of research.

\subsubsection{Projection theory in information geometry}\label{subsec:proj}
To prove Theorem~\ref{th:best_rank1_NMMF}, we first prepare the general projection theory in information geometry~\cite{Amari16}.

The exponential family is a set of probability distributions that can be represented as $\log{P_{\boldsymbol\theta}(x)} = C(x)+\sum^N_{i=1} \theta_i F_i(x) - \psi(\boldsymbol\theta)$ using \emph{natural parameters} $\boldsymbol{\theta}=( \theta_{1}, \dots, \theta_{N}) \in \mathbb{R}^{N}$. It is known that \emph{expectation parameters} $\eta_i=\mathbb{E}_{P_{\boldsymbol\theta}}[F_i]$ can be uniquely determined by the normalization factor $\psi(\boldsymbol{\theta})$ and natural parameters $\boldsymbol{\theta}$.
Since both natural parameters $\boldsymbol{\theta}$ and expectation parameters $\boldsymbol{\eta} = (\eta_1, \dots, \eta_N)$ can identify a distribution, we can use them as coordinate systems for distributions, which is a typical approach in information geometry. %

Let $\mathcal{Q}$ be the set of distributions that satisfy a condition $f(\boldsymbol\theta_{1:n})=0$ for a linear function $f(\cdot)$ on a part of the natural parameters $\boldsymbol{\theta}_{1:n}= (\theta_{1},\dots,\theta_{n})$, where we assume that this part is from $1$ to $n$ without loss of generality.
This set $\mathcal{Q}$ is a subspace characterized by the $\theta$-coordinate system. Let us find a distribution $Q\in\mathcal{Q}$ that is nearest to a given distribution $P$ in terms of the KL divergence; that is, solve the optimization problem
    \begin{align*}
            Q =\argmin_{Q' \in \mathcal{Q}} D(P,Q').
    \end{align*}
This constrained optimization is called the \emph{$m$-projection} of $P$ onto $\mathcal{Q}$. When the condition of $\mathcal{Q}$ is linear to $\boldsymbol{\theta}$, the projection destination $Q$ always uniquely exists and this optimization becomes a convex problem. Moreover, this $m$-projection does not change the rest of the part of expectation parameters $\boldsymbol{\eta}_{n+1:N}=(\eta_{n+1},\dots,\eta_{N} )$~\cite{amari2008information}. In this paper, we call this property \emph{expectation conservation law in $m$-projection}. It has already been reported that we can conduct non-negative tensor Tucker rank reduction efficiently using this conservation law~\cite{ghalamkari2021fast}.

\begin{figure}[t]
\centering
\includegraphics[width=.9\linewidth]{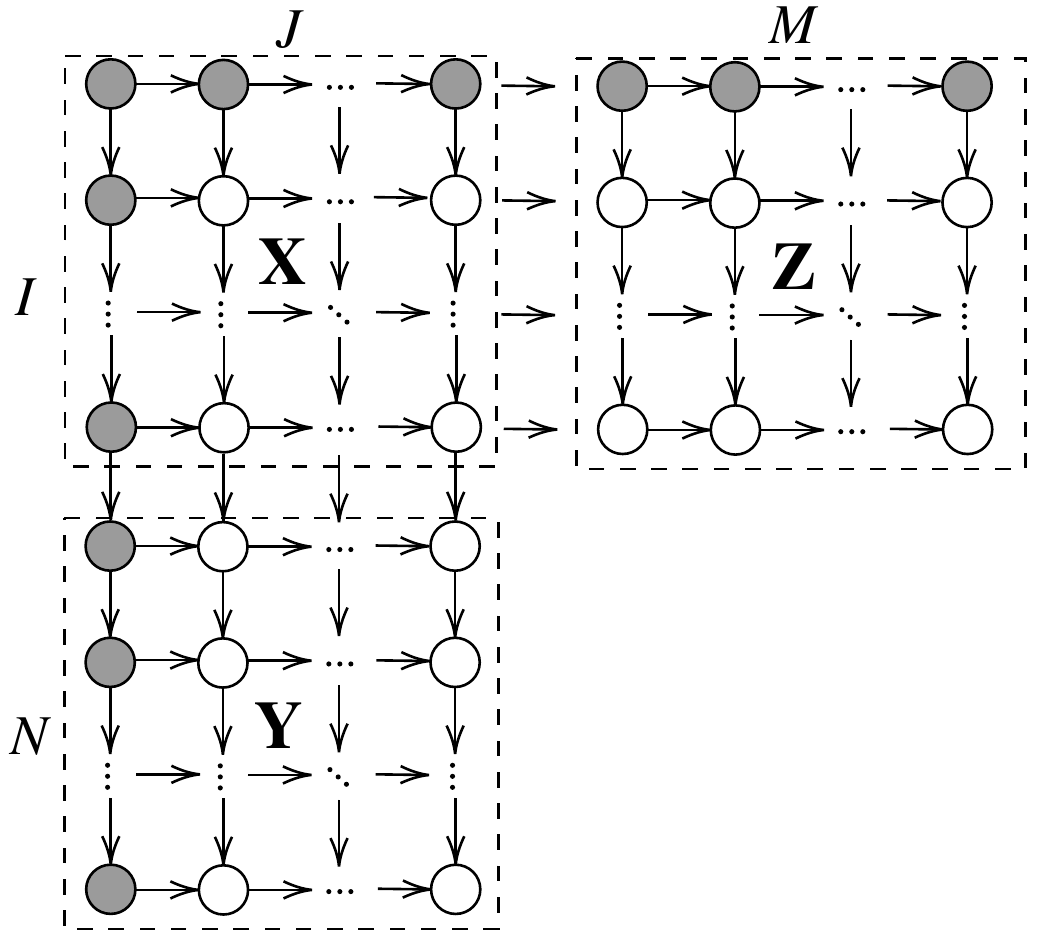}
\caption{A partial order structure for NMMF for three input matrices $\mathbf{X}\in\mathbb{R}_{>0}^{I\times J},\mathbf{Y}\in\mathbb{R}_{>0}^{N \times J}$, and $\mathbf{Z}\in\mathbb{R}_{>0}^{I\times M}$. Only natural parameters on gray-colored nodes can have non-zero values if and only if $(\mathbf{X},\mathbf{Y},\mathbf{Z})$ is simultaneously rank-1 decomposable.}
\label{fig:poset}
\end{figure}

\subsubsection{Modeling}\label{subsec:model}
The input of NMMF is a triple $\left(\mathbf{X},\mathbf{Y},\mathbf{Z}\right)$, where $\mathbf{X} \in \mathbb{R}^{I\times J}_{> 0}$, $\mathbf{Y}\in\mathbb{R}_{> 0}^{N \times J}$, and $\mathbf{Z}\in\mathbb{R}_{> 0}^{I \times M}$. For simplicity, we normalize them beforehand so that their sum is $1$; that is, $S(\mathbf{X})+S(\mathbf{Y})+S(\mathbf{Z})=1$. It is straightforward to eliminate this assumption using the property of KL divergence, $\alpha D\left(\mathbf{X},\mathbf{Y}\right)=D\left(\alpha\mathbf{X},\alpha\mathbf{Y}\right)$, for any non-negative number $\alpha$.
We model these three matrices using a single discrete distribution equipped with a structured sample space.

Let $p$ be a probability mass function with the sample space $\Omega$ given as
\begin{align*}
    \Omega &= \Omega_{\mathbf{X}} \cup \Omega_{\mathbf{Y}} \cup \Omega_{\mathbf{Z}}, \ \text{where}\\
    \Omega_{\mathbf{X}} &= [I] \times [J],\quad
    \Omega_{\mathbf{Y}} = [I + 1, I + N] \times [J],\\
    \Omega_{\mathbf{Z}} &= [I] \times [J + 1, J + M].
\end{align*}
For each $(k, l) \in \Omega$, probability $p(k, l)$ is defined as 
    \begin{align}\label{eq:model}
            p(k,l) = \exp\left(
            \sum_{(s,t) \leq (k,l)} \theta_{st}
            \right),
    \end{align}
where we use a partial order ``$\le$'' over elements of $\Omega$, defined as $(k,l) \leq (s,t)$ if and only if $k \leq s$  and $l \leq t$. It is clear that this distribution in Equation~\eqref{eq:model} belongs to the exponential family and $\boldsymbol{\theta} = (\theta_{ij})_{(i, j) \in \Omega \setminus \{(1,1)\}}$ are natural parameters of the distribution, hence there is a one-to-one correspondence between $(\mathbf{X},\mathbf{Y},\mathbf{Z})$ and $\boldsymbol{\theta}$. 
The natural parameter $\theta_{11}$, in which $(1, 1)$ is the smallest element in the partially ordered space $\Omega$, corresponds to the normalization factor $\psi(\boldsymbol{\theta})$. 
The natural parameters are identified so that they satisfy
    \begin{align*}
        \mathbf{X}_{ij}=p(i,j), \ \  \mathbf{Y}_{nj}=p(I+n,j), \ \  \mathbf{Z}_{im}=p(i,J+m),
    \end{align*}
where the subspace $\Omega_{\mathbf{X}}$ corresponds to $\mathbf{X}$, $\Omega_{\mathbf{Y}}$ to $\mathbf{Y}$, and $\Omega_{\mathbf{Z}}$ to $\mathbf{Z}$.
This model is a special case of a log-linear model on a poset~\cite{Sugiyama17ICML}.
Figure~\ref{fig:poset} illustrates the partial order for the input triple $(\mathbf{X,Y,Z})$. 
There are other possible ways to model $\left(\mathbf{X},\mathbf{Y},\mathbf{Z}\right)$ as a probability distribution using different partial order structure. However, the solution formula that we obtain in Theorem~\ref{th:best_rank1_NMMF} does not depend on the modeling.

Since we use the exponential function in our modeling in Equation~\eqref{eq:model}, our model cannot treat 0. Therefore, our mathematical statements are guaranteed with only positive inputs.

Our model belongs to the exponential family, which makes it possible to analyze NMMF from the viewpoint of information geometry~\cite{Amari16}. The generalization of this model, the log-linear model on posets, is flexible to handle discrete structures, so our model is able to handle the partial order structure of the NMMF.
As described in the previous subsection, we can also identify an exponential distribution by the expectation parameters $\boldsymbol\eta$ instead of the natural parameters $\boldsymbol\theta$. Using results provided in~\cite{Sugiyama17ICML} based on the incidence algebra between natural parameters $\boldsymbol\theta$ and expectation parameters $\boldsymbol\eta$, we can also obtain the expectation parameters $\boldsymbol\eta$ using the following formula:
    \begin{align*}
        \eta_{kl}=\sum_{(k,l)\leq(s,t)}p(s,t).
    \end{align*}
Therefore we have a one-to-one mapping between an input tuple $\left(\mathbf{X},\mathbf{Y},\mathbf{Z}\right)$, $\boldsymbol{\theta}$, and $\boldsymbol\eta$.

\subsubsection{Derivation of the exact solution of rank-1 NMMF}\label{subsec:the_best_NMMF}
 Let $\bm{w}\in\mathbb{R}_{\geq 0}^I, \bm{h}\in\mathbb{R}_{\geq 0}^J, \bm{a}\in\mathbb{R}_{\geq 0}^N$, and $\bm{b}\in\mathbb{R}_{\geq 0}^M$. If three positive matrices $\mathbf{X} \in \mathbb{R}^{I\times J}_{> 0}$, $\mathbf{Y}\in\mathbb{R}_{> 0}^{N \times J}$, and $\mathbf{Z}\in\mathbb{R}_{> 0}^{I \times M}$ can be decomposed into a form $\bm{w}\otimes\bm{h}, \bm{a}\otimes\bm{h}$, and $\bm{w}\otimes\bm{b}$, we say that $\left(\mathbf{X,Y,Z}\right)$ is \emph{simultaneously rank-1 decomposable}.
 Note that the subspace $\mathcal{Q}$ in Section~\ref{subsec:proj} corresponds to the set of simultaneous rank-$1$ decomposable tuples.

\begin{figure}[t]
\centering
\includegraphics[width=.9\linewidth]{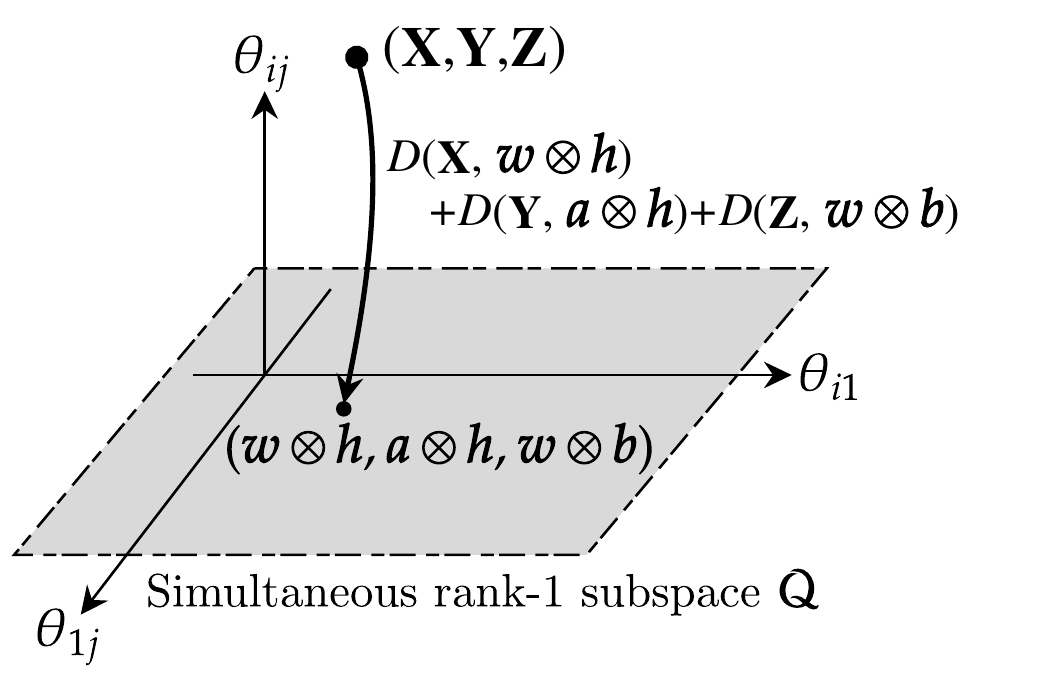}
\caption{Information geometric view of rank-1 NMMF. Rank-1 NMMF is $m$-projection onto simultaneous rank-$1$ subspace from a triple of input three matrices, where one-body expectation parameters do not change.}
\label{fig:IG_NMMF}
\end{figure}

To describe the necessary and sufficient conditions for $\left(\mathbf{X},\mathbf{Y},\mathbf{Z}\right)$ to be simultaneously rank-1 decomposable, we define \emph{one-body} parameters and \emph{two-body} parameters, which were also used by~\citet{ghalamkari2021fast}. A parameter $\theta_{ij}$ or $\eta_{ij}$ is a one-body parameter if one of the indices is $1$; that is, $i=1$ or $j=1$. Parameters other than one-body parameters are two-body parameters; that is, $\theta_{ij},\eta_{ij}$ are two-body parameters if $i\neq1$ and $j\neq1$. Using these parameters, we obtain the following propositions.

\begin{Proposition.}[simultaneous rank-$1$ $\theta$-condition]
     A triple $\left(\mathbf{X,Y,Z}\right)$ is simultaneously rank-1 decomposable if and only if its all two-body natural parameters are $0$; that is, for any $(i,j)\in\Omega$,
     \begin{align*}
         \theta_{ij}=0 \quad \mathrm{if} \ i\neq1 \ \mathrm{and} \ j \neq 1.
     \end{align*}
\end{Proposition.}
In Figure~\ref{fig:poset}, only gray-colored elements have non-zero values of natural parameters $\boldsymbol{\theta}$ if and only if $(\mathbf{X, Y,Z})$ is simultaneously rank-1 decomposable. Moreover, we provide the simultaneously rank-$1$ decomposable condition using $\boldsymbol{\eta}$-parameters as well as on $\boldsymbol{\theta}$-parameters.

\begin{Proposition.}[simultaneous rank-$1$ $\eta$-condition]
     A triple $\left(\mathbf{X,Y,Z}\right)$ is simultaneously rank-1 decomposable if and only if its all two-body expectation parameters are factorizable as a product of two one-body parameters: for any $(i,j)\in\Omega$,
     \begin{align*}
         \eta_{ij}=\eta_{i1}\eta_{1j}.
     \end{align*}
\end{Proposition.}

We call a subspace that satisfies simultaneous rank-$1$ condition \emph{simultaneous rank-$1$ subspace}. From the viewpoint of information geometry, we can understand the best rank-$1$ NMMF as follows (shown in Figure~\ref{fig:IG_NMMF}). The input of NMMF $\left(\mathbf{X},\mathbf{Y},\mathbf{Z}\right)$ corresponds to a point in the space described by the $\boldsymbol{\theta}$-coordinate system. The best rank-$1$ NMMF is an $m$-projection onto the simultaneous rank-$1$ subspace from the input point. 

Since the $m$-projection is a convex optimization, we can get the projection destination by a gradient method. However, it requires appropriate settings for initial values, stopping criterion, and learning rates. In addition, the gradient method is often computationally time-consuming because it involves iterative procedures.

Our closed analytical formula of the projection destination in Theorem~\ref{th:best_rank1_NMMF} solves all the drawbacks of the gradient-based  optimization. According to the expectation conservation law in this $m$-projection onto simultaneous rank-$1$ subspace, two-body $\boldsymbol{\eta}$-parameters do not change in the $m$-projection. That is, for any $(i,j)\in\Omega$,
\begin{align*}
    \eta_{1j} = \overline{\eta}_{1j}, \quad 
    \eta_{i1} = \overline{\eta}_{i1},
\end{align*}
where $\eta$ is the expectation parameter of input, and $\overline\eta$ is the expectation parameter after the $m$-projection onto simultaneous rank-$1$ subspace. By the definition of expectation parameters, we obtain
    \begin{empheq}[left={\empheqlbrace}]{alignat=4}
        {\eta}^{\mathbf{X}}_{1j} - {\eta}^{\mathbf{X}}_{1,j+1} &= \left( S(\bm{w}) + S(\bm{a}) \right)h_j \nonumber \\
        {\eta}^{\mathbf{X}}_{i1} - {\eta}^{\mathbf{X}}_{i+1,1} &= w_i\left( S(\bm{h}) + S(\bm{b}) \right) \nonumber \\
        {\eta}^{\mathbf{Y}}_{1j} - {\eta}^{\mathbf{Y}}_{1,j+1}  &= S(\bm{w})b_j \nonumber \\
        {\eta}^{\mathbf{Z}}_{i1} - {\eta}^{\mathbf{Z}}_{i+1,1} &= a_iS(\bm{h}). \nonumber
    \end{empheq}
where $\eta^{\mathbf{X}}_{ij}=\eta_{ij}$ for $(i,j)\in\Omega_{\mathbf{X}}$, $\eta^{\mathbf{Y}}_{i-I,j}=\eta_{i,j}$ for $(i,j)\in\Omega_{\mathbf{Y}}$, and $\eta^{\mathbf{Z}}_{i,j-J}=\eta_{i,j}$ for $(i,j)\in\Omega_{\mathbf{Z}}$. 

The expectation conservation law guarantees that the the values of the left-hand sides do not change before the $m$-projection, and they also do not change after the $m$-projection. Since the sum of each matrix $\mathbf{X,Y,Z}$ is represented by one-body $\boldsymbol{\eta}$-parameters, the sum of each matrix does not change either in $m$-projection. Using these facts and multiplying these equations together, we can derive Theorem~\ref{th:best_rank1_NMMF}. A complete proof of Theorem~\ref{th:best_rank1_NMMF} is in Supplementary Material.

\begin{figure}[t]
\centering
\includegraphics[width=.99\linewidth]{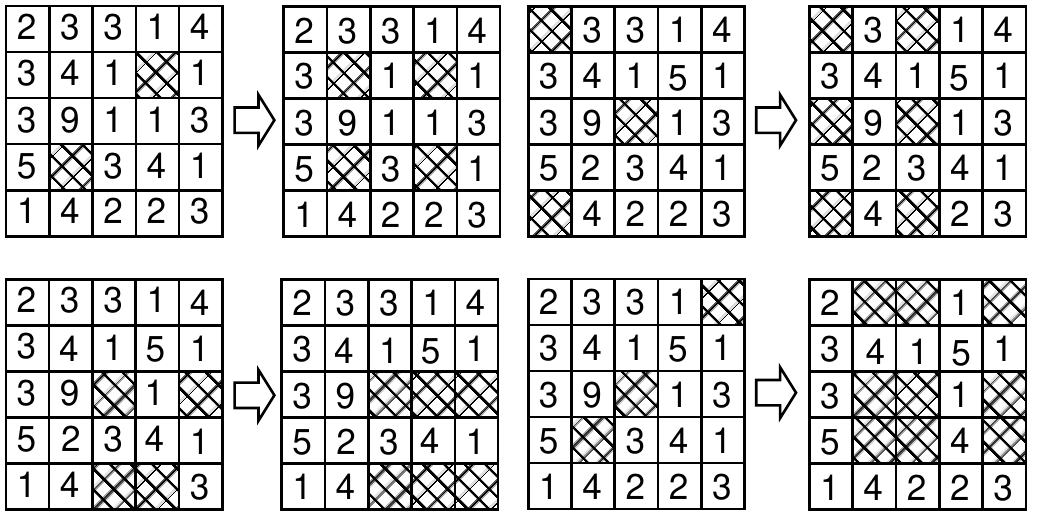}
\caption{Examples of matrices with non-grid-like missing values~(left) and grid-like missing values~(right). Meshed blocks are missing values. We can create grid-like missing values by increase missing values.}
\label{fig:grid}
\end{figure}

\begin{figure*}[t]
\centering
\includegraphics[width=.99\linewidth]{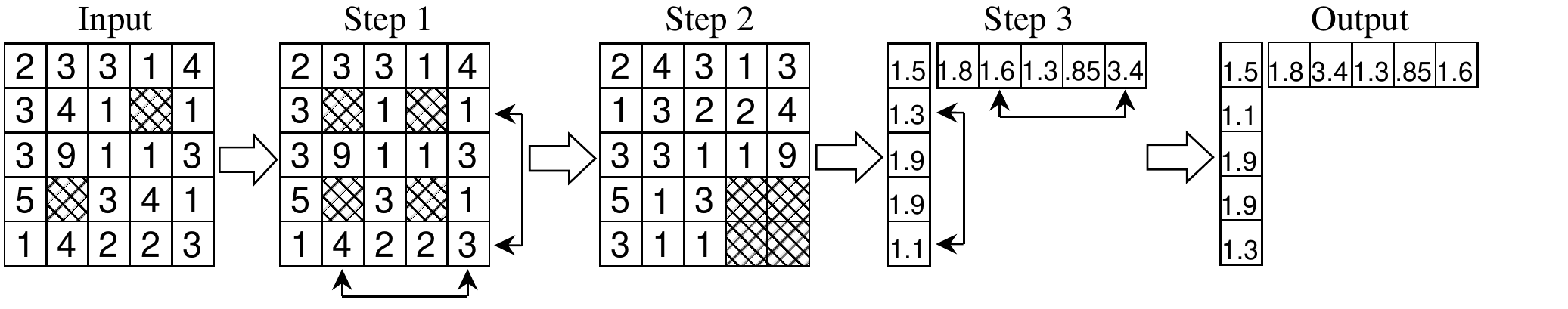}
\caption{Sketch of the algorithm of A1GM. Meshed blocks are missing values. In Step 1, we increase missing values so that they become grid-like. In Step 2, we gather missing values in the block at the bottom right by low and column permutations. In Step 3, we use the closed formula of the best rank-$1$ NMMF in Theorem~\ref{th:best_rank1_NMMF}. In this example, we get $\bm{w}=\left(1.5, 1.3, 1.9\right),\bm{a}=\left(1.9,1.1\right),\bm{h}=\left(1.8,1.6,1.3\right),\bm{a}=\left(0.85,3.4\right)$. Finally, we get two vectors as the output by the repermutation. We use two significant digits in this figure.}
\label{fig:example_A1GM}
\end{figure*}

\subsection{Rank-1 missing NMF using Theorem~\ref{th:best_rank1_NMMF}}\label{subsec:A1GM}

In missing NMF, NMF for missing data, when a binary weight matrix $\mathbf{\Phi}\in\set{0,1}^{I+N,M+J}$ and an input matrix $\mathbf{T}\in\mathbb{R}^{I+N,M+J}_{\geq 0}$ are given in the form of
    \begin{align}\label{eq:zero_right_bottom}
    \mathbf{\Phi} = 
        \begin{bmatrix}
        \mathbf{1}_{IJ} & \mathbf{1}_{IM} \\
        \mathbf{1}_{NJ} & \mathbf{0}_{NM} \\
        \end{bmatrix}, \quad
        \mathbf{T} = 
        \begin{bmatrix}
        \mathbf{X} & \mathbf{Z} \\
        \mathbf{Y} & \mathbf{U} \\
        \end{bmatrix},
    \end{align}
the cost function of missing NMF is equivalent to that of NMMF~\cite{takeuchi2013non}:
    \begin{align*}
        &\argmin_{\mathbf{T}_1;\mathrm{rank}(\mathbf{T}_1)=1} D_\Phi(\mathbf{T},\mathbf{T}_1 )\\
        &= \argmin_{\bm{w,h,a,b}} D(\mathbf{X},\bm{w}\otimes\bm{h})\!+\! D(\mathbf{Y},\bm{a}\otimes\bm{h}) \!+\! D(\mathbf{Z},\bm{w}\otimes\bm{b}).
    \end{align*}
Therefore, using the closed formula of the best rank-$1$ NMMF in Theorem~\ref{th:best_rank1_NMMF}, we can solve the rank-$1$ missing NMF when the binary weight matrix $\mathbf{\Phi}$ is given in the form in Equation~\eqref{eq:zero_right_bottom}. 
In what follows, we always assume that positions of zero entries of $\mathbf{\Phi}$ coincide with those of missing values of a given input matrix $\mathbf{T}$, and we identify them in our discussion.

\subsubsection{A1GM with grid-like missing values}\label{subsec:grid_like}
Using the fact that rank-$1$ missing NMF is homogeneous for row and column permutations, we can develop a method for rank-$1$ missing NMF, called A1GM.

\begin{Proposition.}[Homogeneity of rank-$1$ missing NMF]
    Let $\mathrm{NMF}_1(\mathbf{\Phi},\mathbf{X})$ be the best rank-$1$ matrix $\bm{w}\otimes\bm{h}$, which minimizes the cost function in Equation~\eqref{eq:cost_WNMF}. For any permutation matrices $\mathbf{G}$ and $\mathbf{H}$, it holds that
     \begin{align*}
        \mathrm{NMF}_1\left(\mathbf{G\Phi H},\mathbf{GTH}\right)
        =\mathbf{G}^{\mathrm{T}}\mathrm{NMF}_1\left(\mathbf{\Phi},\mathbf{T}\right)\mathbf{H}^{\mathrm{T}}.
     \end{align*}
\end{Proposition.}

We can then guarantee that A1GM gives the best solution that globally minimizes Equation~\eqref{eq:cost_NMMF} if an input binary weight is grid-like, defined as follows.
    \begin{Definition.}[grid-like binary weight matrix]
        Let $\mathbf{\Phi} \in\{0,1\}^{I+N,J+M}$ be a binary weight matrix. If there exist two sets $S^{(1)} \subset [I+N]$ and $S^{(2)} \subset [J+M]$ such that
            \begin{empheq}[left={\mathbf{\Phi}_{ij}=\empheqlbrace}]{alignat=2}
                0 &\quad \mathrm{ if } \ \  i \in S^{(1)}\ \mathrm{and}\  j \in S^{(2)}, \nonumber \\
                1 &\quad \mathrm{otherwise}, \nonumber
            \end{empheq}
        $\mathbf{\Phi}$ is called grid-like.
    \end{Definition.}
Real-world tabular datasets tend to have missing values on only certain rows or columns. Therefore, the binary weight matrix $\mathbf{\Phi}$ often becomes grid-like in practice (we show examples in Section~\ref{sec:experiments}). Figure~\ref{fig:grid} illustrates examples of matrices with grid-like missing values. 

When $\mathbf{\Phi} \in\{0,1\}^{I+N,J+M}$ is grid-like, we can transform it in the form given in Equation~\eqref{eq:zero_right_bottom} using only row and column permutations.
Let $S^{(1)} \subset [I+N]$ and $S^{(2)} \subset [J+M]$ with $|S^{(1)}| = K$ and $|S^{(2)}| = L$ be the row and column index sets for zero entries in $\mathbf{\Phi}$.
For the block at the bottom right of $\mathbf{\Omega}$ whose row and column indices are specified as
    \begin{align*}
        B^{(1)}&=[I+N-K+1,I+N], \\ 
        B^{(2)}&=[J+M-L+1,J+M],
    \end{align*}
we can collect all the zero entries of $\mathbf{\Phi}$ in the rectangular region $B^{(1)} \times B^{(2)}$ using only row and column permutations. Formally, for a grid-like binary weight matrix $\mathbf{\Phi}$, there are row $\mathcal{G} : [I+N] \to [I+N]$ and column $\mathcal{H}: [J+M] \to [J+M]$ permutations satisfying
    \begin{empheq}[left={(\mathbf{G}\mathbf{\Phi}\mathbf{H})_{ij}=\empheqlbrace}]{alignat=2}
        0 &\quad \mathrm{ if } \ \  i \in B^{(1)} \ \mathrm{and}\ j \in B^{(2)}  \nonumber \\
        1 &\quad \mathrm{otherwise} \nonumber
    \end{empheq}
where $\mathbf{G}$ and $\mathbf{H}$ are corresponding permutation matrices to $\mathcal{G}$ and $\mathcal{H}$, respectively.

We can obtain $\mathbf{G}$ and $\mathbf{H}$ as follows.
First, we focus on row permutation $\mathcal{G}$.
We want to include each row $j\in S^{(1)}\cap B^{(1)c}$, where $B^{(1)c}=[I+N]\setminus B^{(1)}$, in $B^{(1)}$ by row permutation $\mathcal{G}$, which can be achieved by any one-to-one mapping from $S^{(1)}\cap B^{(1)c}$ to $S^{(1)c}\cap B^{(1)}$, where $S^{(1)c}=[I+N]\setminus S^{(1)}$.
Note that $|S^{(1)}\cap B^{(1)c}| = |S^{(1)c}\cap B^{(1)}|$ always holds.
The corresponding permutation matrix is given as
    \begin{align*}
        \mathbf{G} = \prod_{k \in S^{(1)}\cap B^{(1)c}} \mathbf{R}^{k\leftrightarrow \mathcal{G}(k)}
    \end{align*}
where $\mathbf{R}^{k \leftrightarrow l}$ is a permutation matrix, which switches the $k$-th row and the $l$-th row; that is,
    \begin{empheq}[left={\mathbf{R}^{k \leftrightarrow l}_{ij}=\empheqlbrace}]{alignat=2}
      & 0       &\qquad &\text{if $(i,j) = (k,k) \ \mathrm{ or } \ (l,l)$,}  \nonumber\\
      & 1       &       &\text{if $(i,j) = (k,l) \ \mathrm{ or } \ (l,k)$,}  \nonumber\\
      & \mathbf{I}_{ij}  &       &\text{otherwise}.  \nonumber
    \end{empheq}
Since $S^{(1)}\cap B^{(1)^c}$ and $S^{(1)c}\cap B^{(1)}$ are disjoint, it holds that $\mathbf{G}=\mathbf{G}^\mathrm{T}$.

In the same way, any one-to-one mapping from $S^{(2)}\cap B^{(2)c}$ to $S^{(2)c}\cap B^{(2)}$ can be $\mathcal{H}$ and the corresponding permutation matrix is given as 
    \begin{align*}
        \mathbf{H} = \prod_{k \in S^{(2)}\cap B^{(2)c}} \mathbf{R}^{k\leftrightarrow \mathcal{H}(k)},
    \end{align*}
which is also a symmetric matrix.

The above discussion leads to the following procedure of the best rank-1 missing NMF for an input matrix $\mathbf{T}$ if a binary weight matrix $\mathbf{\Phi}$ is grid-like. The first step is to find proper permutations $\mathbf{G}$ and $\mathbf{H}$ to collect the missing values in the lower right corner. In the next step, we obtained $\mathrm{NMF}_1\left(\mathbf{G\Phi H},\mathbf{GTH}\right)$ using the closed formula of the best rank-$1$ NMMF. In the final step, we operated the inverse permutations of $\mathbf{G}$ and $\mathbf{H}$ to the result of the previous step; that is, $\mathbf{G}^{-1}\mathrm{NMF}_1\left(\mathbf{G\Phi H},\mathbf{GTH}\right)\mathbf{H}^{-1}$. Note that $\mathbf{G}^{-1}=\mathbf{G}^\mathrm{T} = \mathbf{G}$ and $\mathbf{H}^{-1}=\mathbf{H}^\mathrm{T} = \mathbf{H}$ always holds since these permeation matrices are orthogonal and symmetrical.

\subsubsection{A1GM for general case}

If a binary weight matrix $\mathbf{\Phi}$ is not grid-like, the above procedure cannot be directly applied.
To treat any matrices with missing values, we \emph{increase} missing values so that $\mathbf{\Phi}$ becomes grid-like. Although this strategy is counter-intuitive because we lose some information, which may cause a larger reconstruction error, we gain the efficiency instead using our closed-form solution in Theorem~\ref{th:best_rank1_NMMF}, and, as we empirically show in the next section, the error increase is not significant in many datasets.
Examples of this step are demonstrated in Figure~\ref{fig:grid}.

In the worst case, the number of missing values after this step becomes $k^2$ for  $k$ missing values. If every row or column has at least one missing value, all indices are missing after this step, for which our algorithm does not work. Thus, our method is not suitable if there are too many missing values in a matrix. 

We illustrate an example of the overall procedure of A1GM in Figure~\ref{fig:example_A1GM} and show its algorithm in Algorithm~\ref{alg:A1GM}.
Since the time complexity of each process of A1GM is at most linear with respect to the number of entries of an input matrix, the time complexity is $O((I+N)(J+M))$.

\paragraph{Relation to \textit{em}-algorithm}
As a method to solve rank-$1$ NMF with missing values, the $em$-algorithm repeats the following $e$- and $m$-steps after filling missing values of an input matrix $\mathbf{T}$ with arbitrary values~\cite{zhang2006learning}.

\begin{inparaenum}
\item[\bf{$m$-step}]: Get the rank-$1$ approximation of the input matrix $\mathbf{T}$ that minimizes the KL divergence from $\mathbf{T}$. \\
\item[\bf{$e$-step}]: Overwrite missing values of $\mathbf{T}$ by the obtained matrix in the $m$-step with keeping other values.
\end{inparaenum}

This algorithm also minimizes the cost function~\eqref{eq:cost_WNMF} indirectly~\cite{Amari16}. For grid-like data, A1GM directly finds the convergence point of the $em$-algorithm without performing the above iterations.

\IncMargin{1em}
\begin{algorithm}[t]\label{alg:A1GM}
\SetKwInOut{Input}{input}
\SetKwInOut{Output}{output}
\SetFuncSty{textrm}
\SetCommentSty{textrm}
\SetKwFunction{BWLRR}{{\scshape A1GM}}
\SetKwProg{myfunc}{}{}{}

\Input{A binary weight matrix $\mathbf{\Phi}\in\set{0,1}^{I \times J}$, a matrix $\mathbf{X}\in\mathbb{R}_{\geq 0}^{I \times J}$}
\Output{Dominant factors $\bm{c}\in\mathbb{R}_{\geq 0}^{I}$ and $\bm{d}\in\mathbb{R}_{\geq 0}^{J}$}

\myfunc{\BWLRR{$\mathbf{\Phi}$, $\mathbf{X}$}}{
$S^{(1)} \leftarrow \emptyset$;
$S^{(2)} \leftarrow \emptyset$\\
\For{$(i,j)\in[I]\times[J]$}{
\If{$ \mathbf{\Phi}_{ij} = 0$}
{
    $S^{(1)} \gets S^{(1)} \cup \{i\}$;
    $S^{(2)} \gets S^{(2)} \cup \{j\}$\\
}}

$B^{(1)} \leftarrow \{I-|S^{(1)}|+1, I-|S^{(1)}|+2, \dots, I\}$ \\
$B^{(2)} \leftarrow \{J-|S^{(2)}|+1, J-|S^{(2)}|+2, \dots, J\}$ \\

$\mathrm{perm1} \leftarrow (1,\dots,I)$ \\
\For{$k \in \{1, 2, \dots, |S^{(1)} \cap B^{(1)c}|\}$}{
    $i \gets$ $k$th smallest element of $S^{(1)} \cap B^{(1)c}$\\
    $j \gets$ $k$th smallest element of $S^{(1)c} \cap B^{(1)}$\\
    swap($\mathrm{perm1}[i]$,$\mathrm{perm1}[j]$) \\
}
$\mathrm{perm2} \leftarrow (1,\dots,J)$ \\
\For{$k \in \{1, 2, \dots, |S^{(2)} \cap B^{(2)c}|\}$}{
    $i \gets$ $k$th smallest element of $S^{(2)} \cap B^{(2)c}$\\
    $j \gets$ $k$th smallest element of $S^{(2)c} \cap B^{(2)}$\\
    swap($\mathrm{perm2}[i]$,$\mathrm{perm2}[j]$) \\
}
$\mathbf{X} \leftarrow  \mathbf{X}[\mathrm{perm1},\mathrm{perm2}]$ \\
$\bm{w},\bm{h},\bm{a},\bm{b}\leftarrow$ the best rank-$1$ NMMF of $\mathbf{X}$ \\
$\bm{c}\leftarrow$ concate $\bm{w}$ and $\bm{a}$;
$\bm{d}\leftarrow$ concate $\bm{h}$ and $\bm{b}$ \\
$\bm{c}\leftarrow \bm{c}[\mathrm{perm}1]$;
$\bm{d}\leftarrow \bm{d}[\mathrm{perm}2]$ \\

\Return{$\bm{c},\bm{d}$}
} 

\caption{A1GM}
\label{alg:LTR}
\end{algorithm}
\section{EXPERIMENTS}\label{sec:experiments}
In this section, we empirically investigate the efficiency and effectiveness of our proposed method, A1GM, on synthetic and real data. We use three types of data: (i) synthetic data with missing values at the lower right corner, (ii) synthetic data with random grid-like missing values, and (iii) real data with grid-like and non-grid-like missing values. It is guaranteed that A1GM always finds the best solution for any data of (i) and (ii). Thus, we only investigate efficiency in our experiments for (i) and (ii). By contrast, for data (iii), the reconstruction error can be worse than the existing methods due to increased missing values in A1GM. Therefore, in the experiment for data (iii), we investigate both efficiency (running time) and effectiveness (reconstruction error). 



We use KL-WNMF as a comparison method~\cite{blondel2007weighted}. KL-WNMF is a commonly used gradient method that reduces the KL-based cost in Equation~\eqref{eq:cost_WNMF} by multiplicative updates. Although faster NMF methods, such as ALS~\cite{kim2014algorithms} and ADMM~\cite{hajinezhad2016nonnegative}, have been developed, they are just as fast as a multiplicative update when the target rank is small~\cite{song2014fast}. Moreover, as we will show in Sections~\ref{subsec:synth} and~\ref{subsec:real}, KL-WNMF converges within only 2--4 iterations in our experiments. In addition, the $em$-algorithm needs more iterations since it minimizes the cost function indirectly. Thus, these techniques are considered to be ineffective for speeding up rank-$1$ KL-WNMF. This is why we only compared A1GM with simple KL-WNMF.

We implemented KL-WNMF by referring to the original paper~\cite{blondel2007weighted}. The stopping criterion of KL-WNMF follows the implementation of the standard NMF in scikit-learn~\cite{scikit-learn}.
The initial values of KL-WNMF are determined by sampling from a uniform distribution from $0$ to $1$. 

All methods are implemented in \texttt{Julia 1.6}. We used \texttt{BenchmarkTools} to measure the running time~\cite{BenchmarkToolsjl2016}. Experiments were conducted on Ubuntu 20.04.1 with a single core of 2.1GHz Intel Xeon CPU Gold 5218 and 128GB of memory.

\begin{figure}[t]
\centering
\includegraphics[width=\linewidth]{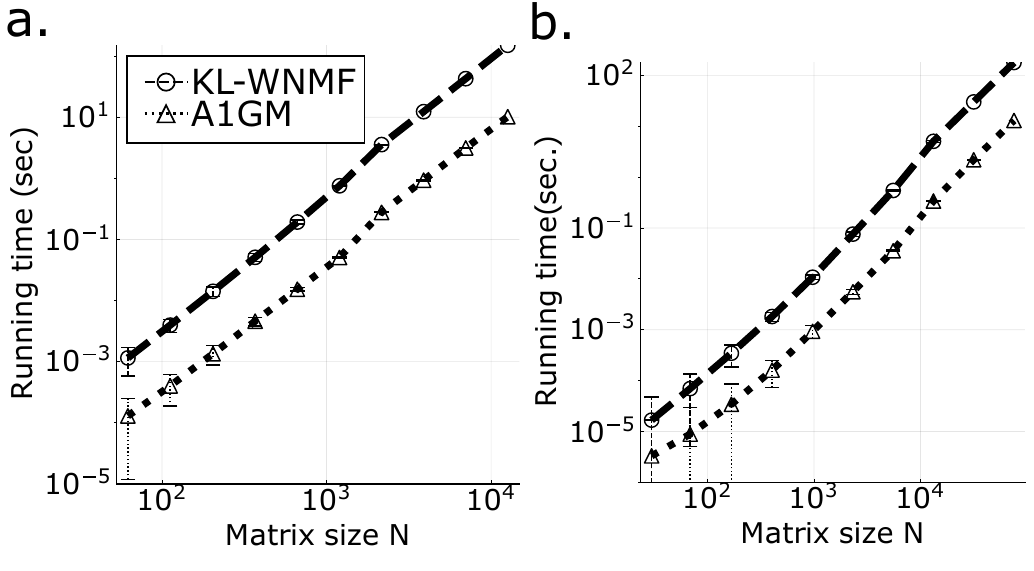}
\caption{Running time comparison of the proposed method A1GM (triangle, dots line) and KL-WNMF (circle, dashed line) with respect to the matrix size $N$. (a) Missing values are at the lower right corner. (b) Missing value positions are grid-like. We plot mean $\pm$ S.D. of five trials.}
\label{fig:exp_synthetic}
\end{figure}

\begin{table*}[t]
\caption{Performance of A1GM compared to KL-WNMF on 17 real datasets.}
\label{tb:performances_A1GM}
\centering
\begin{small}
\begin{tabular}[t]{lcccccccc}
\toprule
DataSet             & size          & \# missing values   & increase rate & relative error & relative runtime \\
\midrule
Autompg             & (398, 8)      & 6             & 1         & 1 & 0.12957 \\
DailySunSpot        & (73718, 9)    & 3247          & 1         & 1 & 0.12845 \\
CaliforniaHousing   & (20640, 9)    & 207           & 1         & 1 & 0.11821 \\
MTSLibrary          & (1533078, 4)  & 1247722       & 1         & 1 & 0.18327 \\
BigMartSaleForecas  & (8522, 5)     & 1463          & 1         & 1 & 0.12699 \\
BoardGameGeekData   & (101375, 17)  & 21            & 1         & 1 & 0.14625 \\
CreditCardApproval  & (590, 7)      & 25            & 1.92      & 1.0018 & 0.12212 \\
HumanResourceAnaly  & (14999, 7)    & 519           & 1.96146   & 1.0168 & 0.11858 \\
heartdisease        & (303, 14)     & 6             & 2         & 1 & 0.12259 \\
lungcancer          & (32, 57)      & 5             & 2         & 1.0001 & 0.13803 \\
PerthHousePrice     & (33656, 14)   & 16585         & 2.61345   & 1.0004 & 0.15382 \\
SleepData           & (62, 8)       & 12            & 2.75      & 1.0211 & 0.18208 \\
arrhythmia          & (452, 280)    & 408           & 4.70588   & 1.0148 & 0.11387 \\
Bostonhousing       & (506, 14)     & 120           & 5.6       & 1.003 & 0.1097 \\
LifeExpectancyData  & (2938, 19)    & 2563          & 7.04097   & 5.7983 & 0.095773 \\
HCCSurvivalDataSet  & (165, 50)     & 826           & 8.3632    & 3.2898 & 0.07113 \\
wiki4HE             & (913, 53)     & 1995          & 18.10175  & 1.2363 & 0.066256 \\

\bottomrule
\end{tabular}
\end{small}

\vskip -0.1in
\end{table*}

\subsection{Synthetic datasets}\label{subsec:synth}
\textbf{Missing values in the lower right corner} 
We prepared synthetic matrices $\mathbf{X}\in\mathbb{R}^{N \times N}$ and their weights $\mathbf{\Phi}\in\set{0,1}^{N \times N}$. We assumed that each input weight $\mathbf{\Phi}$ is in the form of Equation~\eqref{eq:zero_right_bottom}. We measured the running time to obtain rank-$1$ decomposition of $\mathbf{X}$ with varying the matrix size $N$. As Figure~\ref{fig:exp_synthetic}(a) shows, A1GM is an order of magnitude faster than the existing gradient method. The number of iterations of the existing method until convergence was between 2 and 4. A1GM just applies the closed formula in Theorem~\ref{th:best_rank1_NMMF} to parts of input matrices.

\textbf{Random grid-like missing values}
 We also prepared synthetic matrices and its binary weight matrices $\mathbf{\Phi}\in\set{0,1}^{N \times N}$. We assumed that every input weight matrix $\mathbf{\Phi}$ is grid-like and we set the ratio of missing values to be 5 percent. We measured the running time of A1GM to complete the best rank-$1$ missing NMF compared with KL-WNMF by varying the matrix size $N$. As Figure~\ref{fig:exp_synthetic}(b) shows, our method is always faster than the gradient method. The number of iterations of the existing method required for convergence was between 2 and 4. Note that in these datasets A1GM does not need to increase missing values.

\subsection{Real datasets}\label{subsec:real}

We used 17 real datasets. We downloaded tabular datasets that have missing values from the Kaggle databank\footnote{https://www.kaggle.com/datasets} or UCI dataset.\footnote{https://archive.ics.uci.edu/ml/} If a dataset contains negative values, we converted them to their absolute values. Zero values in a matrix were replaced with the average value of the matrix to make them all positive. See Supplementary Material for the sources of these datasets.
We evaluate the relative error as
\begin{align*}
    D_{\mathbf{\Phi}}(\mathbf{X},\mathrm{A1GM}(\mathbf{X})) \big/ D_{\mathbf{\Phi}}(\mathbf{X},\mathrm{WNMF}(\mathbf{X})),
\end{align*}
where $\mathrm{WNMF}(\mathbf{X})$ and $\mathrm{A1GM}(\mathbf{X})$ are the rank-$1$ reconstructed matrices by KL-WNMF and A1GM, respectively, and the binary weight matrix $\mathbf{\Phi}$ indicates locations of missing values of $\mathbf{X}$. We also compared the relative running time of A1GM to KL-WNMF.

The results are summarized in Table~\ref{tb:performances_A1GM}. 
In the table, the column \texttt{increase rate} means the ratio of the number of missing values after addition in A1GM to the original number of missing values. 
If \texttt{increase rate} is $1$, it means that the location of missing values of the dataset is originally grid-like. For such datasets, it is theoretically guaranteed that our method A1GM always provides the best rank-$1$ missing NMF, which minimizes the KL divergence in Equation~\eqref{eq:cost_WNMF}. It is reasonable that the reconstructed matrix by KL-WNMF and that by A1GM are the same since the cost function~\eqref{eq:cost_WNMF} is convex. The number of iterations of the existing method required for convergence was between 2 and 4 for real datasets.

We can see that A1GM is much faster than KL-WNMF for all the datasets. Moreover, the relative error remains low even if missing values of datasets are not grid-like for most of datasets. In some real data, missing values are likely to be biased towards a particular row or column. As a result, they become grid-like by just adding small number of missing values. In these cases, our proposed method can conduct rank-$1$ missing NMF rapidly with competitive errors to KL-WNMF. 
By contrast, a large amount of information is lost after the increasing missing value step for some datasets (large \texttt{increase rates}). As a result, our method is not suitable for obtaining an accurately reconstructed rank-$1$ matrix, even though it is much faster than the existing method.

\section{CONCLUSION}
In this paper, we have derived the closed analytical formula of the best rank-$1$ NMMF.
To obtain this formula, we have used the conservation law in $m$-projection in information geometry by modeling matrices as a log-linear model on a poset.
Using the formula, we have developed a novel method of rank-$1$ NMF for missing data, called A1GM. We have shown that A1GM obtains the best rank-$1$ NMF when missing values are located in a grid-like manner. When the location of missing values is not grid-like, we increase the number of missing values so that they become grid-like, to which again we can use the closed formula of the best rank-$1$ NMMF. We empirically show that A1GM, which is not based on the gradient descent method, is more efficient than the existing gradient method for rank-$1$ missing NMF.

As noted, our method has two main limitations. First, because our modeling uses a log-linear model, we cannot handle zero values in a matrix. Second, the performance of A1GM is not expected to be convincing if there are a huge number of missing values.
NMMF and NMF for missing data have been extended to tensors as NMTF~\cite{takeuchi2013nonT} and WNTF~\cite{ozerov2013weighted}, respectively. Generalization of our study to tensors is an interesting area for future work. 


\section*{Acknowledgement}
This work was supported by JSPS KAKENHI Grant Number JP20J23179 (KG), JP21H03503 (MS), and JST, PRESTO Grant Number JPMJPR1855, Japan (MS).

\bibliography{main}
\appendix

%

%

\onecolumn
\section{Proofs}
    \setcounter{Proposition.}{0}
    \setcounter{Theorem.}{0}
    \begin{Proposition.}[simultaneous rank-$1$ $\theta$-condition]
        A triple $\left(\mathbf{X,Y,Z}\right)$ is simultaneously rank-1 decomposable if and only if its all two-body natural parameters are $0$, that is, for any $(i,j)\in\Omega$,
        \begin{align*}
         \theta_{ij}=0 \quad \mathrm{if} \ i\neq1 \ \mathrm{and} \ j \neq 1.
        \end{align*}
    \end{Proposition.}
    \begin{proof}
    It is clear from the definition that $(\mathbf{X,Y,Z})$ is simultaneously rank-$1$ decomposable if and only if we can represent the corresponding distribution $p$ as $p(i,j)=s(i)t(j)$ for any index $(i,j)\in\Omega$. 
    
    First, we show that we can represent $p(i,j)=s(i)t(j)$ for any index $(i,j)\in\Omega$ if all two-body natural parameters are 0. If all two-body natural parameters are 0, we get a distribution corresponding to $(\mathbf{X,Y,Z})$ as
        \begin{align*}
             p(i,j)
             = \exp{\left(\theta_{11}\right)}
                \exp{\left(\sum^i_{i'=2}\theta_{i'1}\right)}
                \exp{\left(\sum^{j}_{j'=2}\theta_{1j'}\right)}.
         \end{align*}
    The normalization condition $\sum_{ij}p(i,j)=1$ leads to
        \begin{align*}
            \exp{\left(\theta_{11}\right)} 
            = \frac{1}{\sum^{I+N}_{i=1} \exp{\left(\sum^i_{i'=2} \theta_{i'1}\right) }}
              \frac{1}{\sum^{J+M}_{j=1} \exp{\left(\sum^j_{j'=2} \theta_{j'1}\right) }}.
         \end{align*}
         The distribution can then be written as 
        \begin{align*}
             p(i,j)
             &= \frac{
                \exp{\left(\sum^i_{i'=2}\theta_{i'1}\right)}}
                {\sum^{I+N}_{i=1} \exp{\left(\sum^i_{i'=2} \theta_{i'1}\right) }}
                \frac{
                \exp{\left(\sum^{j}_{j'=2}\theta_{1j'}\right)}}
                {\sum^{J+M}_{j=1} \exp{\left(\sum^j_{j'=2} \theta_{j'1}\right)}} \\
            & \equiv s(i)t(j).
        \end{align*}
    Thus, $p(i,j)=s(i)t(j)$ holds.
    
    Next, we show that all two-body natural parameters are 0 if we can represent $p(i,j)=s(i)t(j)$ for any index $(i,j)\in\Omega$. We consider a contraposition of the statement; that is, there is a index  $(i,j)\in\Omega$ that does not hold $p(i,j)=s(i)t(j)$ if a two-body natural parameter is not 0. When $p(i,j)=s(i)t(j)$, it holds that 
        \begin{align}\label{eq:ratio_st}
            \frac{p(i,j)}{p(i,j+1)}=\frac{p(i',j)}{p(i',j+1)}.
        \end{align}
    for any index $(i,j)\in\Omega$. However, for example, if $\theta_{22} \neq 0$, we can immediately confirm Equation~\eqref{eq:ratio_st} is not correct for $i=1,j=1$, and $i'=2$. Therefore, the proposition has been proved.
        
    \end{proof}

    \begin{Proposition.}[simultaneous rank-$1$ $\eta$-condition]
         A triple $\left(\mathbf{X,Y,Z}\right)$ is simultaneously rank-1 decomposable if and only if all of its two-body expectation parameters are factorizable as a product of two one-body parameters: for any $(i,j)\in\Omega$,
         \begin{align*}
             \eta_{ij}=\eta_{i1}\eta_{1j}.
         \end{align*}
    \end{Proposition.}
    \begin{proof}
        As we can see in Proposition 1, $\left(\mathbf{X,Y,Z}\right)$ is simultaneously rank-1 decomposable if and only if the distribution becomes
        \begin{align*}
             p(i,j) = s(i)t(j)
         \end{align*}
        Then,
        \begin{align*}
            {\eta}_{ij} 
            &= \sum_{(i',j')\geq(i,j)}p(i',j') = \sum_{(i',j')\geq(i,j)} s(i')t(j')
            = \sum_{i' \geq i} s(i') \sum_{(i',j')\in\Omega} s(i')t(j') \sum_{j' \geq j} t(j') \\
            &= \sum_{i' \geq i} s(i') \sum_{j' \geq 1} t(j') \sum_{i' \geq 1} s(i') \sum_{j' \geq j} t(j') \\
            &={\eta}_{i1} {\eta}_{1j}.
         \end{align*}
         We used the normalization condition $\sum_{ij}p(i,j)=1$.
    \end{proof}

    \begin{Theorem.}[the best rank-$1$ NMMF]
    For any given three positive matrices $\mathbf{X} \in \mathbb{R}^{I\times J}_{> 0}$, $\mathbf{Y}\in\mathbb{R}_{> 0}^{N \times J}$, and $\mathbf{Z}\in\mathbb{R}_{> 0}^{I \times M}$ and two parameters $\alpha, \beta \ge 0$, four non-negative vectors $\bm{w}\in\mathbb{R}_{\geq 0}^{I}, \bm{h}\in\mathbb{R}_{\geq 0}^{ J}, \bm{a}\in\mathbb{R}_{\geq 0}^{N}$, and $\bm{b}\in\mathbb{R}_{\geq 0}^{M}$ that minimize the cost function in Equation~\rm{(1)} is given as 
         \begin{align*}
            w_i&=\frac{\sqrt{S(\mathbf{X})}}{S(\mathbf{X})+\beta S(\mathbf{Z})}
            \left\{\sum_{j=1}^J\mathbf{X}_{ij}+\sum_{m=1}^M\beta \mathbf{Z}_{im}\right\}, \\
            h_j&=\frac{\sqrt{S(\mathbf{X})}}{S(\mathbf{X})+\alpha S(\mathbf{Y})}
            \left\{\sum_{i=1}^I\mathbf{X}_{ij}+\sum_{n=1}^N\alpha \mathbf{Y}_{nj}\right\}, \\
            a_{n}&=\frac{\sum_{j=1}^J \mathbf{Y}_{nj}}{\sqrt{S(\mathbf{X})}}, \quad 
            b_{m}=\frac{\sum_{i=1}^I \mathbf{Z}_{im}}{\sqrt{S(\mathbf{X})}}.
         \end{align*}
    \end{Theorem.}
    \begin{proof}
    First, we show this theorem with $\alpha=\beta=1$, followed by generalizing the result for any non-negative $\alpha$ and $\beta$. Hereinafter, we use overline for quantities on the simultaneous rank-$1$ subspace; for example, $(\overline{\mathbf{X}},\overline{\mathbf{Y}},\overline{\mathbf{Z}})$ as rank-$1$ matrices sharing factors, $\overline{\eta}$ as expectation parameters for distributions on simultaneous rank-$1$ subspace. We decompose input matrices $\mathbf{X,Y}$ and $\mathbf{Z}$ as $\overline{\mathbf{X}}=\bm{w}\otimes\bm{h}, \overline{\mathbf{Y}}=\bm{a}\otimes\bm{h}$, and $\overline{\mathbf{Z}}=\bm{w}\otimes\bm{b}$, respectively, so that they minimize the cost function
    \begin{align}
        D(\mathbf{X},\bm{w}\otimes\bm{h})+\alpha D(\mathbf{Y},\bm{a}\otimes\bm{h})+\beta D(\mathbf{Z},\bm{w}\otimes\bm{b}).
    \end{align}
    To simplify, we define as follows:
    \begin{align*}
        \eta^{\mathbf{X}}_{ij}=\eta_{ij}    \quad &\mathrm{for} \ \ (i,j)\in\Omega_{\mathbf{X}}, \\
        \eta^{\mathbf{Y}}_{i-I,j}=\eta_{ij} \quad &\mathrm{for} \ \ (i,j)\in\Omega_{\mathbf{Y}}, \\
        \eta^{\mathbf{Z}}_{i,j-J}=\eta_{ij} \quad &\mathrm{for} \ \ (i,j)\in\Omega_{\mathbf{Z}}.
    \end{align*}
    According to the expectation conservation law in this $m$-projection, it holds that
    \begin{align*}
        \eta^{\mathbf{X}}_{ij}&=\overline\eta^{\mathbf{X}}_{ij} \quad 
        \mathrm{for} \  i=1 \ \mathrm{or} \ j=1, \\
        \eta^{\mathbf{Y}}_{nj}&=\overline\eta^{\mathbf{Y}}_{nj} \quad 
        \mathrm{for} \  n=1, \  \\
        \eta^{\mathbf{Z}}_{im}&=\overline\eta^{\mathbf{Z}}_{im} \quad 
        \mathrm{for} \ m=1,
    \end{align*}
    where $(\eta^{\mathbf{X}}, \eta^{\mathbf{Y}}, \eta^{\mathbf{Z}})$ is the expectation parameter of input, and $(\overline\eta^{\mathbf{X}}, \overline\eta^{\mathbf{Y}}, \overline\eta^{\mathbf{Z}})$ is the expectation parameter after the $m$-projection. 
    By the definition of expectation parameters and the conservation law, we obtain
    \begin{empheq}[left={\empheqlbrace}]{alignat=4}
        {\eta}^{\mathbf{X}}_{1j} - {\eta}^{\mathbf{X}}_{1,j+1} &= 
        \left( S(\bm{w}) + S(\bm{a}) \right)h_j, \nonumber \\
        {\eta}^{\mathbf{X}}_{i1} - {\eta}^{\mathbf{X}}_{i+1,1} &= w_i\left( S(\bm{h}) + S(\bm{b}) \right), \nonumber \\
        {\eta}^{\mathbf{Y}}_{1j} - {\eta}^{\mathbf{Y}}_{1,j+1}  &= S(\bm{w})b_j, \nonumber \\
        {\eta}^{\mathbf{Z}}_{i1} - {\eta}^{\mathbf{Z}}_{i+1,1} &= a_iS(\bm{h}). \nonumber
    \end{empheq}
    We multiply these equations together and simplify them, resulting in
    \begin{align*}
        \overline{\mathbf{X}}_{ij} &= h_iw_j = 
        \frac{
        \left({\eta}^{\mathbf{X}}_{i1} - {\eta}^{\mathbf{X}}_{i+1,1}\right)
        \left({\eta}^{\mathbf{X}}_{1j} - {\eta}^{\mathbf{X}}_{1,j+1}\right)}
        { \left( S(\bm{w}) + S(\bm{a}) \right) \left(S(\bm{h}) + S(\bm{b})\right)}, \\
        \overline{\mathbf{Y}}_{ij} &= a_ih_j = 
        \frac{
        \left({\eta}^{\mathbf{Y}}_{i1} - {\eta}^{\mathbf{Y}}_{i+1,1}\right)
        \left({\eta}^{\mathbf{X}}_{1j} - {\eta}^{\mathbf{X}}_{1,j+1}\right)}
        { \left( S(\bm{w}) + S(\bm{a}) \right) S(\bm{h})}, \\
        \overline{\mathbf{Z}}_{ij} &= w_ib_j = 
        \frac{
        \left({\eta}^{\mathbf{X}}_{i1} - {\eta}^{\mathbf{X}}_{i+1,1}\right)
        \left({\eta}^{\mathbf{Z}}_{1j} - {\eta}^{\mathbf{Z}}_{1,j+1}\right)}
        { S(\bm{w})\left( S(\bm{h}) + S(\bm{b}) \right) } .
    \end{align*}
    
    Since the sum of each matrix $\mathbf{X,Y}$ and $\mathbf{Z}$ are represented by conserved quantities, $S(\mathbf{X})=\eta^{\mathbf{X}}_{11}-\eta^{\mathbf{Y}}_{11}-\eta^{\mathbf{Z}}_{11}$, $S(\mathbf{Y})=\eta^{\mathbf{Y}}_{11}$, and $S(\mathbf{Z})=\eta^{\mathbf{Z}}_{11}$, the sum of each matrix also do not change in this projection, that is, $S(\mathbf{X})=S(\overline{\mathbf{X}}),S(\mathbf{Y})=S(\overline{\mathbf{Y}})$, and $S(\mathbf{Z})=S(\overline{\mathbf{Z}})$. As a result, it follows that 
    \begin{align*}
        S(\mathbf{X})&=S(\overline{\mathbf{X}})=S(\bm{w}\otimes\bm{h})=S(\bm{w})S(\bm{h}), \\
        S(\mathbf{Y})&=S(\overline{\mathbf{Y}})=S(\bm{a}\otimes\bm{h})=S(\bm{a})S(\bm{h}), \\
        S(\mathbf{Z})&=S(\overline{\mathbf{Z}})=S(\bm{w}\otimes\bm{b})=S(\bm{w})S(\bm{b}),
    \end{align*}
    and we obtain
    \begin{empheq}[left={\empheqlbrace}]{alignat=4}
        \overline{\mathbf{X}}_{ij} &= \frac{S(\mathbf{X})}{\left( S(\mathbf{X})+S(\mathbf{Y})\right)\left( S(\mathbf{X})+S(\mathbf{Z})\right)}
        \left(\sum_{j'}\mathbf{X}_{ij'}+\sum_{j'} \mathbf{Z}_{ij'}\right)\left(\sum_{i'}\mathbf{X}_{i'j}+\sum_{i'} \mathbf{Y}_{i'j}\right),
        \nonumber \\ 
        \overline{\mathbf{Y}}_{ij} &= 
        \frac{1}{S(\mathbf{X})+S(\mathbf{Y})}\left(\sum_{i'}\mathbf{X}_{i'j}+\sum_{i'} \mathbf{Y}_{i'j}\right)
        \left(\sum_{j'} \mathbf{Y}_{ij'}\right),\nonumber \\
        \overline{\mathbf{Z}}_{ij} &=
        \frac{1}{S(\mathbf{X})+S(\mathbf{Z})}\left(\sum_{j'}\mathbf{X}_{ij'}+\sum_{j'} \mathbf{Z}_{ij'}\right) \left(\sum_{i'} \mathbf{Z}_{i'j}\right).
        \nonumber 
    \end{empheq}
    Using the general property of the KL divergence, $\lambda D(\mathbf{P,Q})=D(\lambda\mathbf{P},\lambda\mathbf{Q})$ for any matrices $\mathbf{P,Q}$ and non-negative number $\lambda$, the above result with general $\alpha$ and $\beta$ is obtained as 
    \begin{empheq}[left={\empheqlbrace}]{alignat=4}
        \overline{\mathbf{X}}_{ij} &= 
        \frac{S(\mathbf{X})}{\left( S(\mathbf{X})+\alpha S(\mathbf{Y})\right)\left( S(\mathbf{X})+\beta S(\mathbf{Z})\right)}
        \left(\sum_{j'}\mathbf{X}_{ij'}+\sum_{j'} \beta\mathbf{Z}_{ij'}\right)\left(\sum_{i'}\mathbf{X}_{i'j}+\sum_{i'} \alpha\mathbf{Y}_{i'j}\right),
        \nonumber \\ 
        \overline{\mathbf{Y}}_{ij} &= 
        \frac{1}{\alpha \left(S(\mathbf{X})+\alpha S(\mathbf{Y})\right)}\left(\sum_{i'}\mathbf{X}_{i'j}+\sum_{i'} \alpha\mathbf{Y}_{i'j}\right)
        \left(\sum_{j'} \alpha\mathbf{Y}_{ij'}\right),\nonumber \\
        \overline{\mathbf{Z}}_{ij} &=
        \frac{1}{\beta \left(S(\mathbf{X})+\beta S(\mathbf{Z})\right)}\left(\sum_{j'}\mathbf{X}_{ij'}+\sum_{j'} \beta\mathbf{Z}_{ij'}\right) \left(\sum_{i'} \beta\mathbf{Z}_{i'j}\right).
        \nonumber 
    \end{empheq}
    Thus, the theorem was proved.
    \end{proof}

\begin{Proposition.}[Homogeneity of rank-$1$ missing NMF]
    Let $\mathrm{NMF}_1(\mathbf{\Phi},\mathbf{X})$ be the best rank-$1$ matrix $\bm{w}\otimes\bm{h}$ which minimizes the cost function in Equation~\rm{(2)}. For any permutation matrices $\mathbf{G}$ and $\mathbf{H}$, it holds that
     \begin{align*}
        \mathrm{NMF}_1\left(\mathbf{G\Phi H},\mathbf{GTH}\right)
        =\mathbf{G}^{\mathrm{T}}\mathrm{NMF}_1\left(\mathbf{\Phi},\mathbf{T}\right)\mathbf{H}^{\mathrm{T}}.
     \end{align*}
\end{Proposition.}
\begin{proof}
    We assume $\mathbf{T}\in\mathbb{R}_{\geq 0}^{n \times m}$ and $\mathbf{\Phi}\in\{0,1\}^{n\times m}$.
    Let $\mathbf{G}\in\set{0,1}^{n\times n}$ and $\mathbf{H}\in\set{0,1}^{m\times m}$ be the permutation matrices corresponding to the mappings $\mathcal{G}$ and $\mathcal{H}$, respectively. This means that, for a given matrix $\mathbf{X}$ and its row and column permutation $\mathbf{X}'=\mathbf{GXH}$, we have $\mathbf{X}'_{g(i),h(j)} = \mathbf{X}_{i,j}$. We can also apply the permutation matrices $\mathbf{G}$ and $\mathbf{H}$ to vectors $\bm{w}\in\mathbb{R}_{\geq 0}^{n}$ and $\bm{h}\in\mathbb{R}_{\geq 0}^{m}$. For $\bm{w}'=\mathbf{G}\bm{w}$ and $\bm{h}'=\mathbf{H}\bm{h}$, it holds that $w'_{g(i)}=w_i$ and $h'_{h(j)}=h_j$. We define $\mathbf{w}^*$ and $\mathbf{h}^*$ as
    \begin{align*}
        \bm{w}^*,\bm{h}^* 
        & \equiv \argmin_{\bm{w},\bm{h}} D_{\mathbf{G\Phi H}}\left(\mathbf{GXH},\bm{w}\otimes\bm{h}\right) \\
        & =\argmin_{\bm{w},\bm{h}}\sum_{i=1}^n\sum_{j=1}^m
            \left(
                \mathbf{\Phi}_{g(i)h(j)}a_ib_j-\mathbf{\Phi}_{g(i)h(j)}\mathbf{X}_{g(i)h(j)}\log{a_ib_j}
            \right).
    \end{align*}
    We replace $\bm{w}$ with $\mathbf{G}\bm{w}$ and $\bm{h}$ with $\mathbf{H}\bm{h}$ and we get
    \begin{align*}
        \mathbf{G}\bm{w}^*,\mathbf{H}\bm{h}^*&
        =\argmin_{\bm{w},\bm{h}}\sum_{i=1}^n\sum_{j=1}^m\left(\mathbf{\Phi}_{g(i)h(j)}a_{g(i)}b_{h(j)}-\mathbf{\Phi}_{g(i)h(j)}\mathbf{X}_{g(i)h(j)}\log{a_{g(i)}b_{h(j)}}\right) \\
        &=\argmin_{\bm{w},\bm{h}}\sum_{g(i)=1}^n\sum_{h(j)=1}^m\left(\mathbf{\Phi}_{g(i)h(j)}a_{g(i)}b_{h(j)}-\mathbf{\Phi}_{g(i)h(j)}\mathbf{X}_{g(i)h(j)}\log{a_{g(i)}b_{h(j)}}\right) \\
        &=\argmin_{\bm{w},\bm{h}}\sum_{i=1}^n\sum_{j=1}^m\left(\mathbf{\Phi}_{ij}a_{i}b_{j}-\mathbf{\Phi}_{ij}\mathbf{X}_{ij}\log{a_{i}b_{j}}\right)  \\
        &= D_{\mathbf{\Phi}}\left(\mathbf{X},\bm{w}\otimes\bm{h}\right).
    \end{align*}
    Thus, it holds that $\bm{w}^* \otimes \bm{h}^* = \mathrm{NMF}_1\left(\mathbf{G \Phi H}, \mathbf{GXH}\right)$ and $\mathbf{G} \left(\bm{w}^* \otimes \bm{h}^*\right)\mathbf{H} = \mathrm{NMF}_1\left(\mathbf{\Phi}, \mathbf{X}\right)$. Therefore, we have
    \begin{align*}
        \mathrm{NMF}_1\left(\mathbf{G\Phi H},\mathbf{GTH}\right)
        =\mathbf{G}^{\mathrm{T}}\mathrm{NMF}_1\left(\mathbf{\Phi},\mathbf{T}\right)\mathbf{H}^{\mathrm{T}}.
     \end{align*}
    We use the fact that permutation matrix is always orthogonal; that is, $\mathbf{G}^{-1}=\mathbf{G}^{\mathrm{T}}$ and $\mathbf{H}^{-1}=\mathbf{H}^{\mathrm{T}}$.
\end{proof}

\section{Details of datasets}
We provide the list of the source of the real datasets in Table~\ref{tb:dataset_details}.

\begin{table}[t]
\caption{Dataset details.}
\label{tb:dataset_details}
\centering
\begin{small}
\begin{tabular}[t]{lccl}
\toprule
dataset & \# zeros & \# negatives & URL   \\
\midrule
Autompg             & 0     & 0         &   \scalebox{0.8}{\url{https://www.kaggle.com/uciml/autompg-dataset}}  \\
DailySunSpot        & 16994 & 3247      &   \scalebox{0.8}{\url{https://www.kaggle.com/abhinand05/daily-sun-spot-data-1818-to-2019}}  \\
CaliforniaHousing   & 0 & 20640         &   \scalebox{0.8}{\url{https://www.kaggle.com/harrywang/housing?select=housing.csv}} \\
MTSLibrary          & 200932 & 0        &   \scalebox{0.8}{\url{https://www.kaggle.com/sharthz23/mts-library?select=interactions.csv}}  \\
BigMartSaleForecas  & 526 & 0           &   \scalebox{0.8}{\url{https://www.kaggle.com/arashnic/big-mart-sale-forecast?select=train.csv}}  \\
BoardGameGeekData   & 520624 & 21       &   \scalebox{0.8}{\url{https://www.kaggle.com/mandshaw/games-0918}}  \\
CreditCardApproval  & 797 & 0           &   \scalebox{0.8}{\url{https://www.kaggle.com/redwuie/credit-card-approval?select=train.csv}} \\
HumanResourceAnaly  & 27510 & 0         &   \scalebox{0.8}{\url{https://www.kaggle.com/cezarschroeder/human-resource-analytics-dataset}} \\
heartdisease        & 1149 & 0          &   \scalebox{0.8}{\url{https://archive.ics.uci.edu/ml/datasets/Heart+Disease}} \\
lungcancer          & 107 & 0           &   \scalebox{0.8}{\url{https://archive.ics.uci.edu/ml/datasets/Lung+Cancer}} \\
PerthHousePrice     & 0 & 33656         &   \scalebox{0.8}{\url{https://www.kaggle.com/syuzai/perth-house-prices}}  \\
SleepData           & 0 & 0             &   \scalebox{0.8}{\url{https://www.kaggle.com/mathurinache/sleep-dataset}}\\
arrhythmia          & 67256 & 14250     &   \scalebox{0.8}{\url{https://archive.ics.uci.edu/ml/datasets/Arrhythmia}}  \\
Bostonhousing       & 812 & 0           &   \scalebox{0.8}{\url{https://www.kaggle.com/altavish/boston-housing-dataset}}  \\
LifeExpectancyData  & 3385 & 0          &   \scalebox{0.8}{\url{https://www.kaggle.com/kumarajarshi/life-expectancy-who}}  \\
HCCSurvivalDataSet  & 2416 & 0          &   \scalebox{0.8}{\url{https://archive.ics.uci.edu/ml/datasets/HCC+Survival}}  \\
wiki4HE             & 1801 & 0          &   \scalebox{0.8}{\url{https://archive.ics.uci.edu/ml/datasets/wiki4HE} } \\

\bottomrule
\end{tabular}
\end{small}

\vskip -0.1in
\end{table}
\end{document}